\documentclass[10pt,twocolumn,letterpaper]{article}

\usepackage{cvpr}              

\definecolor{cvprblue}{rgb}{0.21,0.49,0.74}
\usepackage[pagebackref,breaklinks,colorlinks,allcolors=cvprblue]{hyperref}
\usepackage{ntheorem}
\usepackage{booktabs}
\usepackage{rotating}
\usepackage{multirow}
\usepackage{tablefootnote}
\usepackage{multicol}
\usepackage{bm}
\usepackage[accsupp]{axessibility}

\def\paperID{*****} 
\def\confName{MORSE}
\def\confYear{2025}

\title{Towards Efficient Benchmarking of Foundation Models in Remote Sensing: \\%
A Capabilities Encoding Approach}

\author{Pierre Adorni$^{1}$, Minh-Tan Pham$^{1}$, Stéphane May$^{2}$, Sébastien Lefèvre$^{1,3}$\\ \\
$^{1}$ IRISA, Université Bretagne Sud, UMR 6074, Vannes, France\\
$^{2}$ Centre National d’Études Spatiales (CNES), Toulouse, France \\
$^{3}$ UiT The Arctic University of Norway, Tromsø, Norway\\
{\tt\small \{pierre.adorni,minh-tan.pham,sebastien.lefevre\}@irisa.fr, stephane.may@cnes.fr}
}
\newtheorem*{pb*}{Problem}

\begin{document}
\maketitle

\begin{abstract}

Foundation models constitute a significant advancement in computer vision: after a single, albeit costly, training phase, they can address a wide array of tasks. In the field of Earth observation, over 75 remote sensing vision foundation models have been developed in the past four years. However, none has consistently outperformed the others across all available downstream tasks. To facilitate their comparison, we propose a cost-effective method for predicting a model's performance on multiple downstream tasks without the need for fine-tuning on each one. This method is based on what we call ``capabilities encoding.'' The utility of this novel approach is twofold: we demonstrate its potential to simplify the selection of a foundation model for a given new task, and we employ it to offer a fresh perspective on the existing literature, suggesting avenues for future research. Codes are available at \href{https://github.com/pierreadorni/capabilities-encoding}{https://github.com/pierreadorni/capabilities-encoding}
\end{abstract}

\section{Introduction}
\label{sec:intro}
Foundation models (FMs) have transformed deep learning, demonstrating remarkable success in domains such as natural language processing (NLP) and computer vision. This new type of model, trained once on a very large dataset, can then be fine-tuned cost-effectively to address a wide range of so-called downstream tasks. The concept of vision foundation models is particularly relevant to Earth observation (EO), as remote sensing sensors generate vast amounts of new unlabeled data daily. Self-supervised learning (SSL) techniques, often employed by FM designers, can leverage such data. Additionally, the scarcity of quality-annotated data creates a need for robust pretrained models that can be adapted with minimal examples. Recently, the EO community has undertaken significant efforts to adapt the technology of FMs to the specifics of remote sensing data, leading to the development of over 75 Remote Sensing Foundation Models (RSFMs), each differing in backbone architecture, pre-training strategy, and pre-training dataset.

Despite these advances, no RSFM has demonstrated universal superiority across all downstream tasks so far. Instead, we observe high performance variability, influenced by factors such as the domain gap between data used in pre-training and downstream tasks, as well as the suitability of the model architecture for a given task. This variability highlights the importance of in-depth model comparisons to identify their areas of expertise, strengths, and weaknesses. Some recent studies have evaluated RSFMs on a set of diverse downstream tasks \cite{sunRingMoRemoteSensing2023, reedScaleMAEScaleAwareMasked2023, guoSkySenseMultiModalRemote2024}, comparing new methods with existing ones. However, the choice of tasks varies from one study to another, making direct comparisons incomplete. To address this issue, standardized third-party benchmarks have been introduced such as PhiEO Bench \cite{fibaekPhilEOBenchEvaluating2024a}, GEO-Bench \cite{lacosteGEOBenchFoundationModels2023} and PANGAEA \cite{marsocciPANGAEAGlobalInclusive2024},
ensuring that all models are tested on the same set of downstream datasets carefully selected for diversity. Nevertheless, due to the inherently limited number of tasks and datasets included in each benchmark, they cannot fully represent the broad range of challenges and opportunities in EO.

A simple but computationally expensive approach to comprehensively evaluate a new model’s capability would be to fine-tune it on a very large number of publicly available datasets covering various downstream tasks. While this approach provides the most complete assessment of its performance, it requires significant computational resources. In this paper, we propose a method to efficiently estimate model performance across numerous tasks without the need for extensive fine-tuning. Our approach is based on what we term `capabilities encoding'. Our contribution distinguishes itself from other benchmarks in three main ways:
\begin{itemize}
    \item We assemble fine tuning results of many models on many datasets into a common, unified database to facilitate comparisons.
    \item We review the usage of different downstream datasets for comparing RSFMs and provide insights for future works.
    \item We develop a method to estimate the performance of a model on many tasks without comprehensive fine-tuning.
    \item We use the latent space of our prediction method to propose a visual exploration of the literature, to identify trends and help future research.
\end{itemize}
In the next section, we first provide an overview of RSFMs and downstream datasets. Sec. \ref{sec:cap_enc} then describes our method for capacity encoding and embedding. Sec. \ref{sec:exp} provides details about our experiments with extensive analysis and visualizations. Sec. \ref{sec:discussion} provides some discussions and highlights key takeaways.

\section{Background}
\label{sec:background}
We reviewed 36 papers on Remote Sensing Vision Foundation Models from the \textit{Awesome Remote Sensing Foundation Models} repository \cite{guoSkySenseMultiModalRemote2024} and extracted the fine-tuning results of each model on the downstream datasets tested. When authors also fine-tuned other RSFMs or trained fully supervised baselines for comparison, we included these results as well. We frequently observed significant variations in results for the same method (e.g., ResNet50) on the same downstream dataset across different publications. This variability can be attributed to subtle differences in implementation, hyperparameter optimization efforts, and other factors. To address these discrepancies, we selected the highest reported value whenever duplicates were encountered.

We compiled a table of $(\text{model}, \text{task})$ pairs, where each task is represented by a triplet $(\text{dataset, fraction, metric})$. For instance, $(\text{SkySense}, [\text{Potsdam, 100\%, mF1}])$ denotes the mF1 score yielded by the SkySense model \cite{guoSkySenseMultiModalRemote2024} on the Potsdam dataset \cite{sherrahFullyConvolutionalNetworks2016}, using 100\% of the training set for fine-tuning. Each dataset may have multiple variations based on the metric used for comparison, the fraction of the training set employed, and other evaluation criteria (e.g., Potsdam appears as seven different tasks). Additionally, each model may appear multiple times in our data if the authors trained various versions, typically with different backbone sizes. For example, RingMo \cite{sunRingMoRemoteSensing2023} is available in two versions: one with a ViT-B backbone and another with a Swin-B backbone. In the end, we collected 1106 performance values of 85 models on 160 downstream datasets.

\subsection{Remote Sensing Vision Foundation Models}

The initial efforts to develop Remote Sensing Foundation Models (RSFMs) began in 2020, SSL techniques such as MoCo \cite{heMomentumContrastUnsupervised2020} and BYOL \cite{grillBootstrapYourOwn2020}. These techniques were employed to pre-train Convolutional Neural Networks (CNNs) before fine-tuning them on remote sensing vision datasets. Since then, a variety of training strategies and pre-training datasets have been explored.

\begin{figure}
    \centering
    \includegraphics[width=1\linewidth]{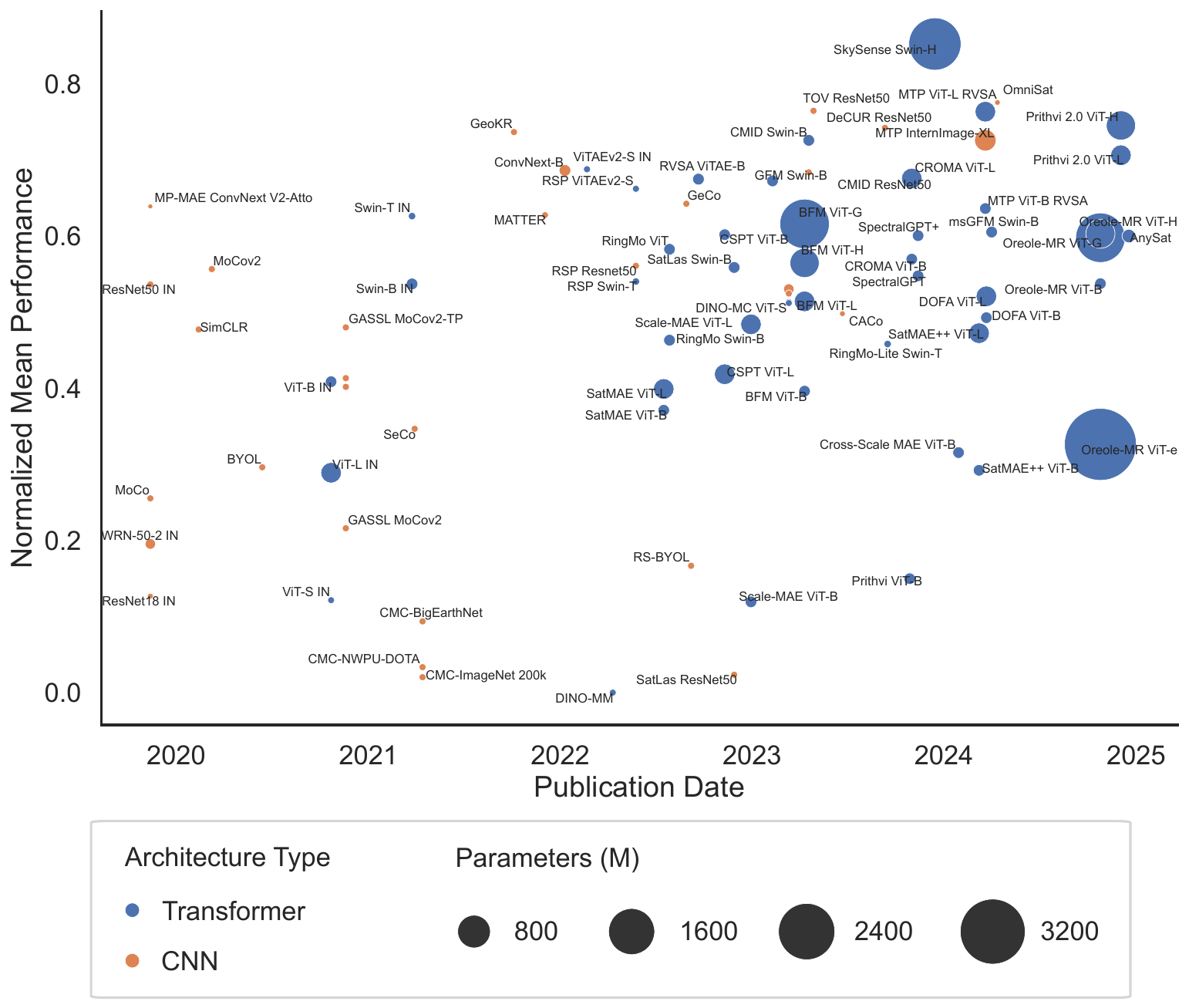}
    \caption{An overview of the main RSFMs. The normalized performance is the position in the literature of a model: its value is 1 when the model is the SOTA on the given task, and 0 when it has the worst performance of the literature. See \cref{sec:norm} for more details.}
    \label{fig:models_overview}
\end{figure}

Today, most RSFMs are large vision transformers, such as ViT\cite{dosovitskiyImageWorth16x162020} or Swin\cite{liuSwinTransformerHierarchical2021}, pre-trained using generative pretext tasks like Masked Autoencoders (MAE)\cite{heMaskedAutoencodersAre2022}. The scalability of transformers has made them an excellent platform for developing foundation models, which may explain their widespread adoption in recent years, as illustrated in \cref{fig:models_overview}. However, the same figure also indicates that while scale is correlated with performance, some smaller networks, such as compact versions of vision transformers or CNNs, can still perform comparably well to their larger transformer-based counterparts. This bodes well for the development of ``pocket'' foundation models and edge computing applications.



\begin{figure*}
    \centering
    \includegraphics[width=\linewidth]{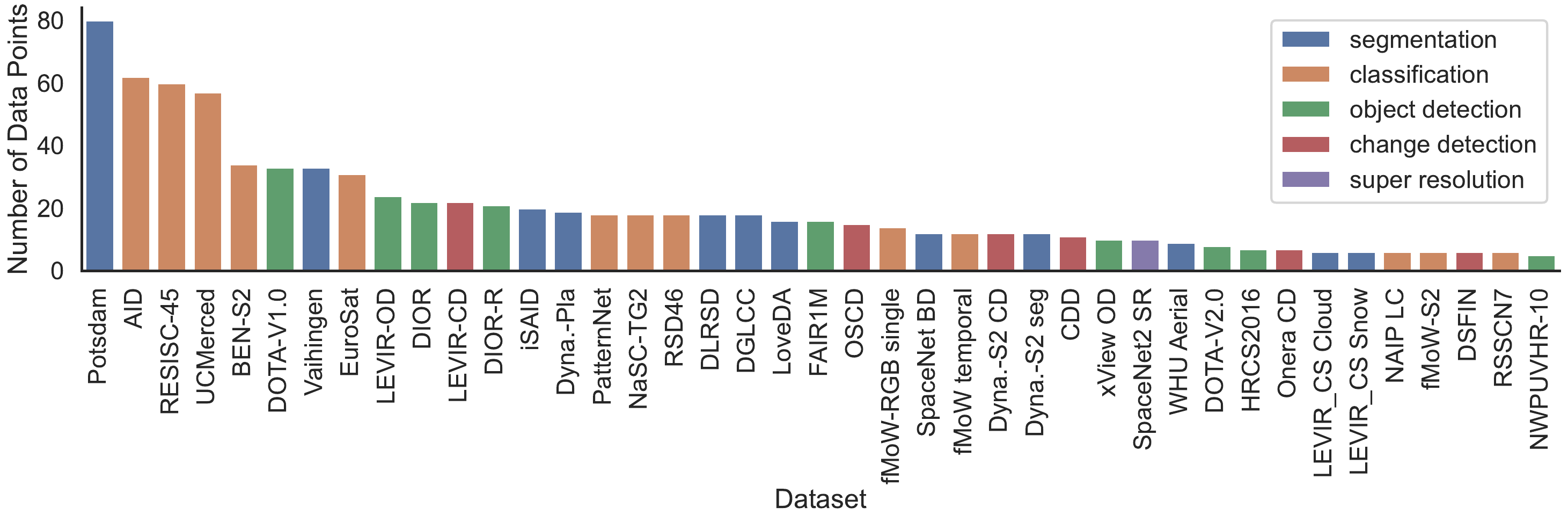}
    \caption{Usage of remote sensing datasets for benchmarking RSFMs in the literature. Most datasets are segmentation or classification datasets, some relate to object detection and change detection, while only one super-resolution dataset is used for benchmarking RSFMs.}
    \label{fig:datasets}
\end{figure*}

\subsubsection*{Pretraining Schemes for RSFMs}

Pretraining strategies for RSFMs leverage both general computer vision techniques and remote-sensing-specific data. Many approaches build on standard computer vision paradigms such as joint embedding methods (e.g., CLIP, DINO) and reconstructive methods (e.g., MAE), while others adapt them to exploit the unique properties of remote sensing data.

One common adaptation is the use of contrastive learning, where positive and negative sample pairs are defined using metadata such as time and location. Methods like SeCo \cite{manasSeasonalContrastUnsupervised2021}, CaCo \cite{mallChangeAwareSamplingContrastive2023}, GASSL \cite{ayushGeographyawareSelfsupervisedLearning2021a}, SatCLIP \cite{klemmerSatCLIPGlobalGeneralPurpose2024}, and SatMIPS \cite{bourcierLearningRepresentationsSatellite2025} follow this paradigm, ensuring that learned representations capture meaningful spatio-temporal relationships. Contrastive learning is also applied across different sensors and modalities --such as optical RGB, multispectral, and SAR imagery-- enabling models to develop cross-modal representations, as seen in msGFM \cite{hanBridgingRemoteSensors2024}, CROMA \cite{fullerCROMARemoteSensing2023}, and DINO-MM \cite{wangSelfsupervisedVisionTransformers2022}.

Other methods introduce task-specific pretraining objectives. For example, GASSL \cite{ayushGeographyawareSelfsupervisedLearning2021a} incorporates a geo-location classification task, effectively using a supervised signal within a self-supervised framework. More broadly, some works adopt fully supervised pretraining strategies, similar to ImageNet, by leveraging large labeled datasets to train models across multiple tasks such as classification, segmentation, and object detection. Notable examples include Satlas \cite{bastaniSatlasPretrainLargeScaleDataset2023} and MTP \cite{wangMTPAdvancingRemote2024}.

Recent approaches increasingly favor hybrid strategies, combining MAE-based reconstruction with contrastive learning objectives to improve representation quality. Additionally, some studies exploit invariances specific to remote sensing, such as learning correspondences between images of the same location captured at different resolutions, as seen in Cross-Scale MAE \cite{tangCrossScaleMAETale}. Overall, modern RSFM pretraining methods integrate multiple strategies, blending contrastive learning, reconstruction, and supervision while leveraging remote sensing-specific characteristics to enhance generalization and robustness.

\subsection{Remote Sensing Downstream Tasks}

\subsubsection{Dataset diversity}

Among the 82 datasets identified in the literature as downstream data for evaluating RSFMs shown in \cref{table:datasets}, 31 are dedicated to segmentation, 30 to classification, 11 to object detection, 9 to change detection, and only 1 to super-resolution task. However, focusing on the top 41 datasets displayed in \cref{fig:datasets}, we find a more balanced distribution: 13 for classification, 12 for segmentation, 9 for object detection, 6 for change detection, and exceptionally 1 for super-resolution. Given that a primary goal of developing foundation models is to create models adaptable to a wide range of downstream tasks, we recommend diversifying the tasks in RSFM benchmarks.

\begin{table*}
\centering
\resizebox*{!}{0.98\textheight}{
    \begin{tabular}{lllr}
    \toprule \textbf{Dataset} & \textbf{Sensor} & \textbf{Best Model} & \textbf{Best Perf.}\\ \midrule
        \multicolumn{4}{l}{\textbf{Classification}} \\ \midrule
        BEN-MM \cite{sumbulBigEarthNetMMLargeScaleMultimodal2021} & Sentinel 1, Sentinel 2 & DeCUR ResNet50 & 89.80 \\ \rowcolor[gray]{0.925}
        PASTIS-HD classif \cite{astrucOmniSatSelfSupervisedModality2024} & Spot 6/7 & AnySat & 72.80 \\ 
        fMoW-S2 \cite{congSatMAEPretrainingTransformers2022} & Sentinel 2 & SkySense Swin-H & 64.38 \\ 
        BEN-S2 \cite{sumbulBigearthnetLargeScaleBenchmark2019} & Sentinel 2 & msGFM Swin-B & 92.90 \\ \rowcolor[gray]{0.925}
        RESISC-45 \cite{chengRemoteSensingImage2017} & Google Earth & DOFA ViT-L & 97.80 \\ 
        RSD46 \cite{xiaoHighResolutionRemoteSensing2017} & Google Earth, Tianditu & TOV ResNet50 & 52.96 \\ \rowcolor[gray]{0.925}
        AID \cite{xiaAIDBenchmarkData2017} & Google Earth & SkySense Swin-H & 97.68 \\ 
        PatternNet \cite{zhouPatternNetBenchmarkDataset2018} & Google Earth & TOV ResNet50 & 85.38 \\ \rowcolor[gray]{0.925}
        UCMerced \cite{neumannIndomainRepresentationLearning2019} & Aerial & RSP ViTAEv2-S & 99.90 \\ 
        RSSCN7 \cite{zouDeepLearningBased2015} & Google Earth & GeoKR & 94.43 \\ \rowcolor[gray]{0.925}
        MLRSNet \cite{qiMLRSNetMultilabelHigh2020} & GeoEye-1, WV-1,2, SPOT-7, Pleiades & ResNet50 IN & 96.43 \\ 
        BigEarthNet-19 \cite{sumbulBigearthnetLargeScaleBenchmark2019} & Sentinel 2 & ResNet50 IN & 81.62 \\ \rowcolor[gray]{0.925}
        fMoW-RGB single \cite{christieFunctionalMapWorld2018} \ & QuickBird-2, GeoEye-1, WV-2, WV-3 & Scale-MAE ViT-L & 77.90 \\ 
        NAIP LC & Aerial & SatMAE ViT-L & 71.77 \\ \rowcolor[gray]{0.925}
        So2Sat \cite{zhuSo2SatLCZ42Benchmark2020} & Sentinel 1, Sentinel 2 & ResNet50 IN & 60.84 \\ 
        fMoW temporal \cite{christieFunctionalMapWorld2018} & QuickBird-2, GeoEye-1, WV-2, WV-3 & SatMAE ViT-L & 81.49 \\ \rowcolor[gray]{0.925}
        EuroSat \cite{helberEuroSATNovelDataset2019} & Sentinel 2 & CROMA ViT-L & 99.46 \\ 
        NaSC-TG2 \cite{zhouNaSCTG2NaturalScene2021} & Tiangong-2 & ResNet50 IN & 95.47 \\ \rowcolor[gray]{0.925}
        Canadian Cropland \cite{jacquesCanadianCroplandDataset2023} & Sentinel 2 & CROMA ViT-L & 78.07 \\ 
        m-bigearthnet \cite{sumbulBigearthnetLargeScaleBenchmark2019,lacosteGEOBenchFoundationModels2023} & Sentinel 2 & MP-MAE ConvNext V2-Atto & 67.10 \\ \rowcolor[gray]{0.925}
        m-forestnet \cite{irvinForestNetClassifyingDrivers2020,lacosteGEOBenchFoundationModels2023} & Landsat-8  & Swin-T IN & 58.00 \\ 
        m-brick-kiln \cite{leeScalableDeepLearning2021, lacosteGEOBenchFoundationModels2023} & Sentinel 2 & ConvNext-B & 98.90 \\ \rowcolor[gray]{0.925}
        m-pv4ger \cite{mayer3DPVLocatorLargescaleDetection2022, lacosteGEOBenchFoundationModels2023} & Google Earth & Swin-T IN & 98.00 \\ 
        m-so2sat \cite{zhuSo2SatLCZ42Benchmark2020, lacosteGEOBenchFoundationModels2023} & Sentinel 1, Sentinel 2 & ConvNext-B & 58.10 \\ \rowcolor[gray]{0.925}
        TreeSatAI \cite{ahlswedeTreeSatAIBenchmarkArchive2023} & Sentinel 1, Sentinel 2 & DOFA ViT-L & 71.60 \\ 
        TreeSatAI-TS \cite{ahlswedeTreeSatAIBenchmarkArchive2023} & Sentinel 1, Sentinel 2 & AnySat & 75.10 \\ \rowcolor[gray]{0.925}
        m-eurosat \cite{helberEuroSATNovelDataset2019, lacosteGEOBenchFoundationModels2023} & Sentinel 2 & ConvNext-B & 97.70 \\ 
        Sen4Map land cover \cite{sharmaSen4MapAdvancingMapping2024} & Sentinel 2 & Prithvi 2.0 ViT-H & 76.10 \\ \rowcolor[gray]{0.925}
        Sen4Map crop type \cite{sharmaSen4MapAdvancingMapping2024} & Sentinel 2 & Prithvi 2.0 ViT-H & 84.60 \\   \midrule

        \multicolumn{4}{l}{\textbf{Segmentation}} \\ \midrule
        FLAIR \cite{garioudFLAIRCountryScaleLand2023a} & Aerial & DOFA ViT-L & 74.90 \\ \rowcolor[gray]{0.925}
        Dyna.-Pla \cite{tokerDynamicEarthNetDailyMultiSpectral2022} & Planet Fusion & CACo & 51.29 \\ \rowcolor[gray]{0.925}
        Dyna.-S2 seg \cite{tokerDynamicEarthNetDailyMultiSpectral2022} & Sentinel 2 & SkySense Swin-H & 46.20 \\ 
        ISPRS -- Potsdam \cite{sherrahFullyConvolutionalNetworks2016} & Aerial & SkySense Swin-H & 93.99 \\ \rowcolor[gray]{0.925}
        iSAID \cite{waqaszamirISAIDLargescaleDataset2019} & Gaofen, JL-1, Google Earth, CycloMedia & SkySense Swin-H & 70.91 \\ 
        ISPRS -- Vaihingen \cite{sherrahFullyConvolutionalNetworks2016} & Aerial & CMID Swin-B & 80.01 \\ \rowcolor[gray]{0.925}
        DGLCC \cite{demirDeepGlobe2018Challenge2018} & WV-3, WV-2, GeoEye-1 & SeCo & 46.40 \\ 
        DLRSD \cite{shaoPerformanceEvaluationSingleLabel2018} & Aerial & SeCo & 43.77 \\ \rowcolor[gray]{0.925}
        LEVIR\_CS Cloud \cite{chenSpatialTemporalAttentionBasedMethod2020} & Gaofen-1 & GeCo & 77.65 \\ 
        LEVIR\_CS Snow \cite{chenSpatialTemporalAttentionBasedMethod2020} & Gaofen-1 & GeCo & 57.78 \\ \rowcolor[gray]{0.925}
        SpaceNet BD \cite{ettenSpaceNetRemoteSensing2019} & PlanetLab Dove & MATTER & 81.12 \\ 
        Sen12MS \cite{schmittSEN12MSCURATEDDATASET2019} & Sentinel 1, Sentinel 2 & RS-BYOL & 82.00 \\ \rowcolor[gray]{0.925}
        LoveDA \cite{wangLoveDARemoteSensing2021} & Spaceborne & BFM ViT-G & 54.40 \\ 
        GID \cite{tongLandcoverClassificationHighresolution2020}& Gaofen-2 & CSPT ViT-B & 64.97 \\ \rowcolor[gray]{0.925}
        WHU Aerial \cite{jiFullyConvolutionalNetworks2019} & Aerial & MTP InternImage-XL & 95.59 \\ 
        Inria \cite{maggioriCanSemanticLabeling2017} & Aerial & Scale-MAE ViT-L & 84.20 \\ \rowcolor[gray]{0.925}
        SpaceNet2 Shanghai \cite{ettenSpaceNetRemoteSensing2019} & WV-3 & Scale-MAE ViT-L & 82.20 \\ 
        SpaceNet2 Vegas \cite{ettenSpaceNetRemoteSensing2019} & WV-3 & Scale-MAE ViT-L & 87.40 \\ \rowcolor[gray]{0.925}
        SpaceNet2 Paris \cite{ettenSpaceNetRemoteSensing2019} & WV-3 & Scale-MAE ViT-L & 81.10 \\ 
        SpaceNet2 Khartoum \cite{ettenSpaceNetRemoteSensing2019} & WV-3 & Scale-MAE ViT-L & 77.10 \\ \rowcolor[gray]{0.925}
        DFC2020 \cite{robinsonGlobalLandCoverMapping2021} & Sentinel 1, Sentinel 2 & CROMA ViT-L & 49.78 \\ 
        DW-Expert \cite{fullerCROMARemoteSensing2023} & Sentinel 1, Sentinel 2 & CROMA ViT-L & 58.71 \\ \rowcolor[gray]{0.925}
        MARIDA \cite{kikakiMARIDABenchmarkMarine2022} & Sentinel 2 & CROMA ViT-B & 65.56 \\ 
        GeoNRW RGB \cite{baierGeoNRW2020} & Aerial, TerraSAr-X & DeCUR ResNet50 & 46.70 \\ \rowcolor[gray]{0.925}
        GeoNRW RGB-DEM \cite{baierGeoNRW2020} & Aerial, TerraSAR-X, Lidar DSM & DeCUR ResNet50 & 48.90 \\ 
        Sen1Floods11 \cite{bonafiliaSen1Floods11GeoreferencedDataset2020} & Sentinel 1 & AnySat & 91.10 \\ \rowcolor[gray]{0.925}
        HLSBurnScars \cite{jakubikFoundationModelsGeneralist2023} & Sentinel 2, Landsat & Prithvi 2.0 ViT-H & 90.50 \\ 
        SegMunich \cite{hongSpectralGPTSpectralRemote2024} & Sentinel 2 & DOFA ViT-B & 51.60 \\ \rowcolor[gray]{0.925}
        m-cashew-plantation \cite{yinMappingSmallholderCashew2023, lacosteGEOBenchFoundationModels2023} &  Sentinel 2 & MP-MAE ConvNext V2-Atto & 79.80 \\ 
        m-SA-crop-type \cite{lacosteGEOBenchFoundationModels2023} & Sentinel 2 & MP-MAE ConvNext V2-Atto & 38.20 \\ \rowcolor[gray]{0.925}
        Landslide4sense \cite{ghorbanzadehLandslide4SenseReferenceBenchmark2022} & Sentinel 2 & Prithvi 2.0 ViT-H & 71.30 \\  \midrule 

        \multicolumn{4}{l}{\textbf{Object Detection}} \\ \midrule
        FAIR1M \cite{sunFAIR1MBenchmarkDataset2022} & Gaofen, Google Earth & SkySense Swin-H & 54.57 \\ \rowcolor[gray]{0.925}
        DIOR-R \cite{liObjectDetectionOptical2020} & Google Earth & MTP ViT-L RVSA & 74.54 \\ 
        DIOR \cite{liObjectDetectionOptical2020} & Google Earth & MTP ViT-L RVSA & 81.10 \\ \rowcolor[gray]{0.925}
        LEVIR \cite{zouRandomAccessMemories2018} & Google Earth & GeCo & 79.72 \\ 
        xView OD \cite{lamXViewObjectsContext2018} & WV-3, WV-2, GeoEye-1 & MTP ViT-L RVSA & 19.40 \\ \rowcolor[gray]{0.925}
        DOTA-V1.0\cite{dingObjectDetectionAerial2022} & Gaofen, JL-1, Google Earth, CycloMedia & MTP ViT-L RVSA & 81.66 \\ 
        DOTA-V2.0 \cite{dingObjectDetectionAerial2022} & Gaofen, JL-1, Google Earth, CycloMedia & BFM ViT-G & 58.69 \\ \rowcolor[gray]{0.925}
        HRSC2016 \cite{liuHighResolutionOptical2017} & Google Earth & ResNet50 IN & 90.40 \\ 
        NWPUVHR-10 \cite{chengMulticlassGeospatialObject2014} & Google Earth, Aerial  & Oreole-MR ViT-H & 89.73 \\ \rowcolor[gray]{0.925}
        UCAS-AOD \cite{zhuOrientationRobustObject2015} & Google Earth & CSPT ViT-B & 90.30 \\ 
        \midrule
        \multicolumn{4}{l}{\textbf{Change Detection}} \\ \midrule
        OSCD \cite{daudtUrbanChangeDetection2018} & Sentinel 2 & SkySense Swin-H & 60.06 \\ \rowcolor[gray]{0.925}
        Dyna.-S2 CD \cite{tokerDynamicEarthNetDailyMultiSpectral2022} & Sentinel 2 & SkySense Swin-H & 18.00 \\ 
        LEVIR-CD \cite{chenSpatialTemporalAttentionBasedMethod2020} & Google Earth & MTP ViT-L RVSA & 92.67 \\ \rowcolor[gray]{0.925}
        CDD \cite{lebedevCHANGEDETECTIONREMOTE2018} & Google Earth & MTP InternImage-XL & 98.33 \\ \rowcolor[gray]{0.925}
        CaiRoad \cite{mallChangeEventDataset2022} & Sentinel 2 & CACo & 44.38 \\ 
        CalFire \cite{mallChangeEventDataset2022}  & Sentinel 2 & CACo & 65.71 \\ \rowcolor[gray]{0.925}
        DSIFN \cite{zhangDeeplySupervisedImage2020} & Google Earth & GFM Swin-B & 71.24 \\  \midrule
        
        \multicolumn{4}{l}{\textbf{Super Resolution}} \\ \midrule
        SpaceNet2 SR \cite{ettenSpaceNetRemoteSensing2019} & WV-3 & ViT-B IN & 23.29 \\ 

        \bottomrule
    \end{tabular}
    }
    \caption{List of the downstream datasets included in our study. The performance metrics used are Overall Accuracy for classification, mf1 or mIoU for segmentation, mAP for object detection, mf1 for change detection and PSNR for super resolution}
    \label{table:datasets}
    
\end{table*}

We observe from \cref{table:datasets} that most public datasets were created using images captured by the Sentinel-2 satellite thanks to its free access and high revisit time. Others were built using high-resolution aerial or Google Earth images, in particular for fine-grained tasks like object detection. Efforts have been made to release very high-resolution optical datasets from commercial satellites such as Pleiades, Worldview-2/3 or Quickbird, but their number is still limited. One can highlight that the literature significantly lacks datasets created from Synthetic Aperture Radar (SAR) and hyperspectral sensors. Finally, thanks again to the availability of Sentinel images, Satellite Image Time Series (SITS) data are becoming increasingly available for benchmarking SITS tasks such as change detection.

\subsubsection{Dataset Quality}

To estimate the quality of a dataset, many options have been explored in previous studies such as visualizing embeddings of the images through a pretrained model as done in \cite{castillo-navarroSemisupervisedSemanticSegmentation2022a}. We choose to estimate the challenge posed by a dataset with the diversity of results in the literature: our hypothesis is that if a task related to a given dataset is easily solved, most methods will yield similar results, and fine-tuning a RSFM on this dataset will not give us very useful information -- it will be considered of low quality. On the contrary, if different methods in the literature perform very differently on a given dataset, fine-tuning a RSFM on it will give us a lot of information on the model's capabilities: it is then considered of high quality. The mean performance is also an indicator of the quality of a dataset: the more room for improvement there is, the better. For example, \cref{tab:dataset_quality} reports the standard deviation ($\sigma$) and mean ($\mu$) of the finetuning results calculated for the most used datasets related to different tasks. 

\begin{table}[h]
\footnotesize
    \centering
    \begin{tabular}{llr}\toprule
        \textbf{Dataset} & \bm{$\sigma$} & \bm{$\mu$} \\ \midrule
        \multicolumn{3}{l}{\textbf{Segmentation}} \\ \midrule
        Potsdam & 4.72 & 83.76 \\
        Vaihingen & 2.87 & 76.19 \\
        iSAID & 2.77 & 65.27 \\ \midrule
        \multicolumn{3}{l}{\textbf{Classification}} \\ \midrule
        AID & 1.41 & 96.95 \\ 
        RESISC-45 & 2.98 & 94.57 \\ 
        UCMerced & 0.6 & 99.19 \\ \midrule
        \multicolumn{3}{l}{\textbf{Object detection}} \\ \midrule
        DOTA-V1.0 & 3.36 & 77.33 \\
        LEVIR & 3.85 & 74.00 \\
        DIOR & 4.16 & 73.13 \\ \midrule
        \multicolumn{3}{l}{\textbf{Change detection}} \\ \midrule
        LEVIR-CD & 2.35 & 90.54 \\
        OSCD & 6.88 & 50.43 \\
        CDD & 1.09 & 96.55 \\ 
        \bottomrule
    \end{tabular}
    \caption{Standard deviation ($\sigma$) and mean ($\mu$) of finetuning results in the literature on the three most used datasets for each task. The metrics used are mIoU for segmentation, overall accuracy for classification, mAP for object detection and F1 for change detection. }
    \label{tab:dataset_quality}
\end{table}

\begin{figure*}
    \centering
    \includegraphics[width=\linewidth]{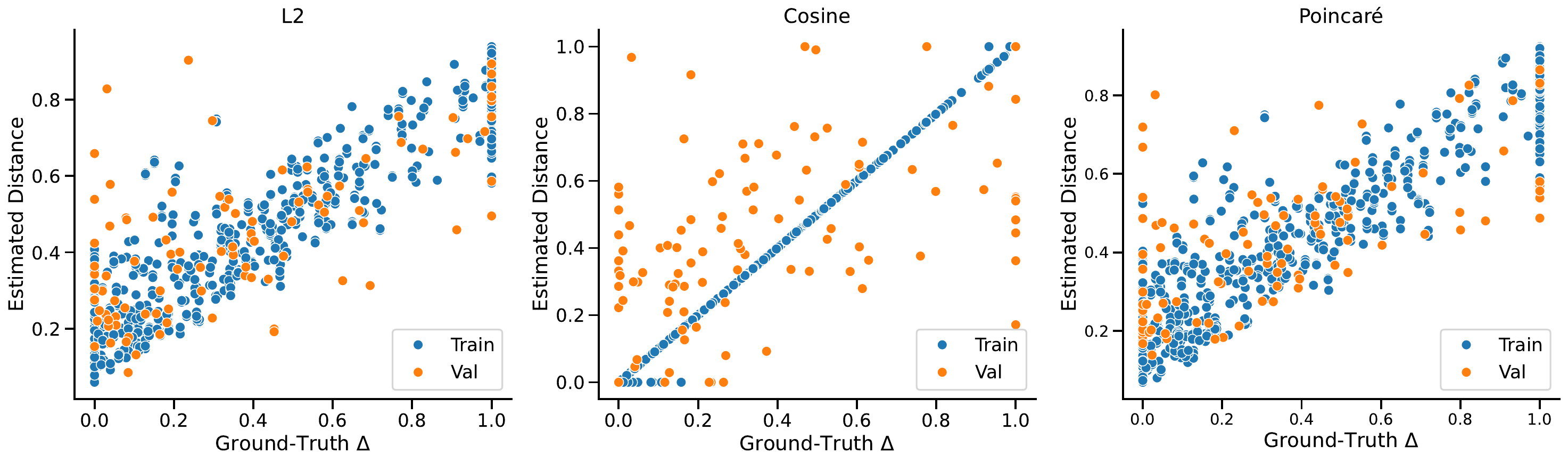}
    \caption{Estimated distance in latent space versus ground truth value $\Delta$ for different distance metrics. L2 and Poincaré metrics perform better than Cosine.}
    \label{fig:metrics_comparison}
\end{figure*}

Several benchmark datasets are saturated: the three most used classification datasets including AID \cite{xiaAIDBenchmarkData2017}, RESISC-45 \cite{chengRemoteSensingImage2017}, and UCMerced \cite{neumannIndomainRepresentationLearning2019} have their mean accuracy approaching $100\%$. Similarly, the Potsdam and Vaihingen segmentation datasets \cite{sherrahFullyConvolutionalNetworks2016} exhibit very high mean performance, coupled with low variance among methods, particularly when considering metrics easier to saturate than mIoU, such as the F1 score. Therefore, we recommend focusing on less-saturated datasets, such as DIOR \cite{liObjectDetectionOptical2020} for object detection and OSCD \cite{daudtUrbanChangeDetection2018} for change detection, and moving away from nearly solved problems to effectively benchmark RSFMs in the future.

Another interesting direction that could be explored more is the settings with very low training labels, also coined few-shot. For example, the RESISC-45 dataset \cite{chengRemoteSensingImage2017} goes from a standard deviation of 2.98 to 7.7 and a mean accuracy of 94.57 to 70 when only 20 samples per class are allowed for training. More generally, generalizing from few examples is a hard problem that can challenge even large RSFMs on datasets that otherwise would be considered easy and somewhat lacking in interest.

\section{Capabilities Encoding}
\label{sec:cap_enc}

\subsection{Normalization}
\label{sec:norm}
To facilitate subsequent processing, we normalize the data. The normalization procedure for each task $t$ is as follows:
\begin{enumerate}
\item Identify the model $m^*$ with the best performance on $t$. The performance of a model $m$ on a task $t$ is noted as $p_{m,t}$, which could be accuracy, F1 score, mIoU, or any other metric where \textit{higher is better}.
\item Compute for each model $m$ the difference $\delta_{m,t} = p_{m^*, t} - p_{m, t}$: the absolute difference from the best performance in the literature for that task.
\item Compute the max-normalized distance for each model $m$: $\Delta_{m,t} = \frac{\delta_{m,t}}{\max\limits_{m\in M}(\delta_{m,t})}$.
\end{enumerate}

An example of $\delta$ and $\Delta$ calculations is shown in~\cref{table:norm}.

\begin{table}[h]
\centering
\begin{tabular}{lcll}
\multicolumn{1}{l}{Model} & \multicolumn{1}{l}{\text{[Potsdam, 100\%, mF1]}} & \multicolumn{1}{l}{$\delta$} & \multicolumn{1}{l}{$\Delta$} \\ \hline
ResNet50 IN                  & 89.7                                    & 4.4                           & 1.0                             \\
\textbf{SkySense}            & \textbf{94.1}                           & \textbf{0.0}                    & \textbf{0.0}                  \\
RingMo                       & 93.2                                    & 0.9                           & 0.2                          
\end{tabular}
\caption{Example of $\delta$ and $\Delta$ calculations in our normalization technique.
}
\label{table:norm}
\end{table}

Thus, we obtain a matrix associating models and tasks with scalars between 0 and 1, where 0 indicates state-of-the-art performance and 1 indicates the worst performance in the literature.

\subsection{Definition}

We pose the following problem:
\begin{pb*}
Is there a set of points $x\in\mathbb{R}^n$, one for each model and task, such that $\forall t,m,\ d(x_t, x_m) = \Delta_{m,t}$ for a given distance $d$? In other words, can we arrange these points so that the distances reflect the values of $\Delta$?
\end{pb*}

Its relaxed version is:
\begin{pb*}
Which set of points minimizes $\sum|\epsilon_{m,t}|$ such that $\forall t,m,\ d(x_t, x_m) + \epsilon_{m,t} = \Delta_{m,t}$?
\end{pb*}

We approximate the solution by optimizing the positions of the points $x_i\in\mathbb{R}^n$ to minimize the MSE loss function:
\begin{equation}
    L(x) = \frac{1}{n}\sum_{\forall t,\forall m} (d(x_m, x_t) - \Delta_{m,t})^2,
\label{eq:loss}
\end{equation}
where $x_t$ and $x_m$ represent the points associated with tasks and models, and $n$ is the total number of points. 



\section{Experiments}
\label{sec:exp}
\subsection{Choice of Embedding Space}

Given that our data is normalized, we can utilize any embedding space that contains values within the range [0,1]. We experimented with fitting our embedding space to $\Delta$ using various distance metrics: Euclidean (L2), Cosine distance, and Poincaré distance (hyperbolic space). In each experiment, a subset of data points was reserved and excluded from the optimization process, subsequently introduced into the embedding space to serve as validation data. For each distance metric, we performed 10 splits: 810 training points and 10 validation points, optimizing the latent space using the 810 training data points.

For each ground truth value $\Delta_{m,t}$, we calculated the distance in the latent space between the corresponding points $p_m$ and $p_t$. In \cref{fig:metrics_comparison}, we plot the computed distance in the embedding space against the ground truth $\Delta$ value for all splits, for each of the three distance metrics. Several observations can be made. The cosine distance exhibits significant overfitting, as the validation data points do not accurately reflect $\Delta$. The L2 and Poincaré distances yield better results, with L2 showing a slight advantage in accuracy. Due to its simplicity and superior performance, we opted to use the L2 distance for subsequent analyses. Further experimentation revealed that a dimension of 5 is optimal, as it provides sufficient complexity for the optimization to converge without allowing overfitting.

\subsection{Exploring the errors}

While our method generally approximates performance well, there remains a high variance, particularly for certain points where predictions significantly deviate from the ground truth value. We investigated potential reasons for these discrepancies. Our primary hypothesis centers on the number of fine-tunings for a model: since our method relies on multiple fine-tuning results from the same model across various datasets to position it within the embedding space, we hypothesized that models with more available fine-tuning results would have more precise positions in the embedding space. This, in turn, would enhance the prediction accuracy for novel datasets.

To address this, we refined our dataset by excluding any model fine-tuned on fewer than five datasets and any dataset used for testing fewer than five models. This data cleaning step reduced our dataset to 820 fine-tuning results of 45 models on 92 datasets, compared to 1,106 fine-tuning results of 85 models on 160 datasets previously.

\begin{figure}
    \centering
    \includegraphics[width=0.95\linewidth]{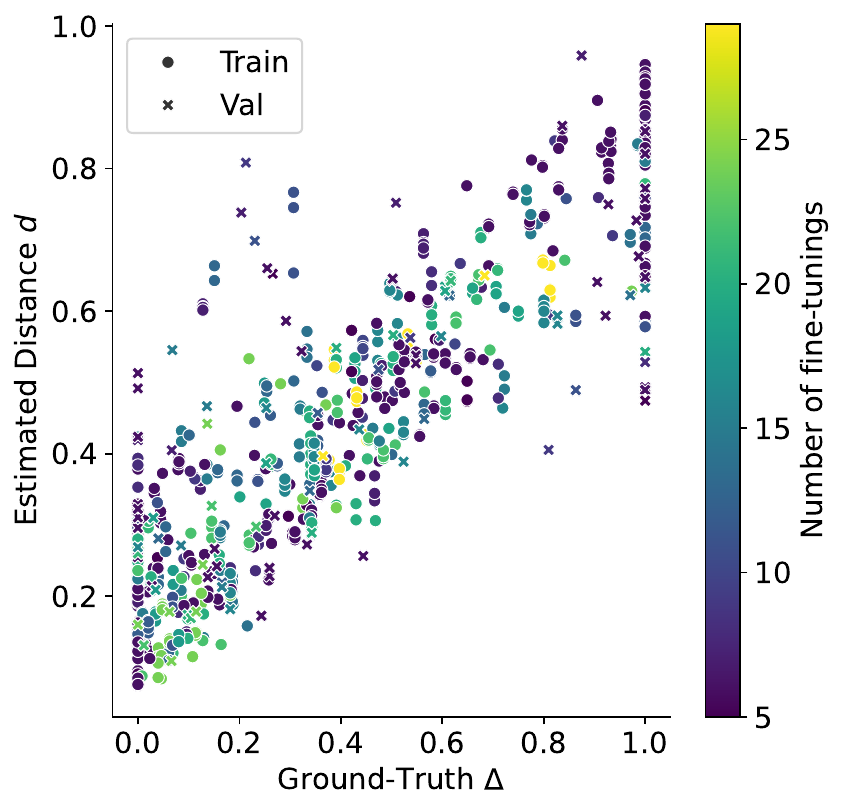}
    \caption{Effect of the number of fine-tunings of a model in the training data on the prediction accuracy.}
    \label{fig:error}
\end{figure}

We then examined the relationship between the number of fine-tunings for a given model and the prediction accuracy for the same model on new datasets. In \cref{fig:error}, each dot represents a fine-tuning result. We observe that models with more extensive testing tend to lie closer to the identity line, exhibiting less variance.

\subsection{Latent Space Visualization}

\begin{figure*}
    \centering
    \includegraphics[width=\linewidth]{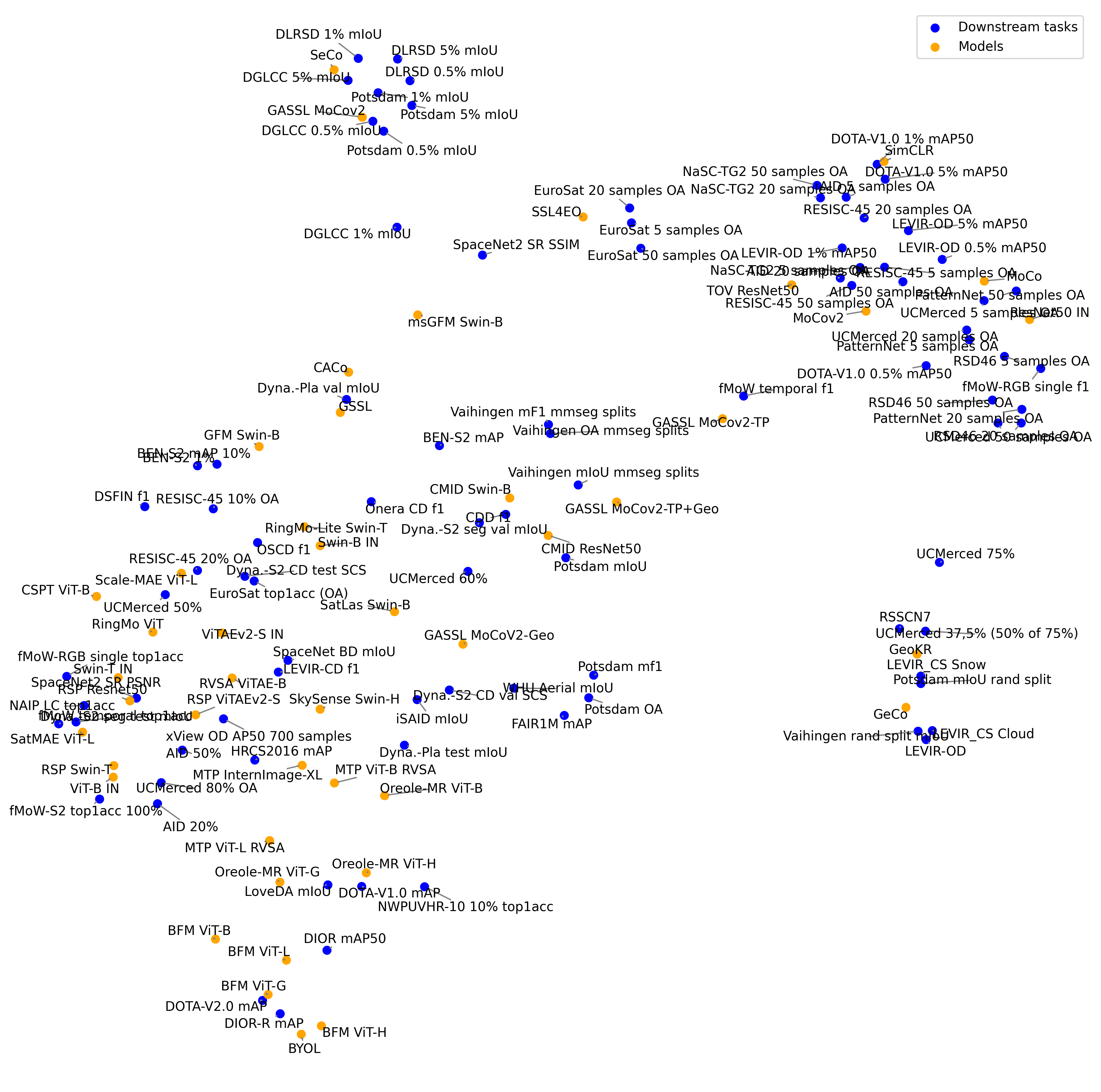}
    \caption{2-dimensional UMAP visualization of the 5-dimensional embedding space. Multiple clusters can be seen, e.g., the few-shot datasets and ResNet based RSFMs on the top right and top left, the main cluster with most methods on the bottom left, etc.}
    \label{fig:latent}
\end{figure*}

We present a visualization of the 5-dimensional embedding space as a 2-dimensional UMAP projection in \cref{fig:latent}. This comprehensive overview of the literature on RSFMs can be utilized to identify emerging trends and clusters of models or datasets, thereby facilitating a deeper understanding of promising avenues for future research. The embedding space reveals multiple clusters: notably, a prominent cluster of few-shot datasets is located in the top right, comprising examples such as the UCMerced classification dataset \cite{neumannIndomainRepresentationLearning2019} with 20 images per class, the LEVIR object detection dataset \cite{zouRandomAccessMemories2018} with only 5\% of the labels, and the DOTA-V1.0 dataset \cite{dingObjectDetectionAerial2022} with only 0.5\% of the labels. An additional smaller few-shot cluster is discernible in the top left of the figure. In both scenarios, all models within these clusters are with CNN architectures, specifically ResNets. This observation implies that CNNs may outperform transformers under conditions of limited image availability. It is noteworthy that the fmMoW RGB\cite{christieFunctionalMapWorld2018} classification dataset is also embedded in the largest few-shot cluster, despite encompassing over a million samples. This anomaly may be attributable to its status as an early large-scale remote sensing dataset, frequently leveraged as a pretraining dataset by early RSFMs, which, notably, predominantly employed CNN architectures.

In the main cluster spanning from the bottom left to the center of the image, datasets and model types are quite diverse. This is because of the way we built the embedding space, i.e., by minimizing distances between points, models which show better performance on many datasets end up in a position of centrality. For example, SkySense \cite{guoSkySenseMultiModalRemote2024}, the best model on most datasets, lies in the center, whereas models like CSPT \cite{zhangConsecutivePreTrainingKnowledge2022}, GASSL \cite{ayushGeographyawareSelfsupervisedLearning2021a} and BFM \cite{chaBillionscaleFoundationModel2024}, which have strengths and weaknesses, are positioned outside of the main cluster. This notion of model centrality can be used as a measure of the generalization capabilities of their embeddings, or the \textit{completeness} of their abilities.

\section{Discussion}
\label{sec:discussion}




One of the primary goals of the method was to predict the accuracy of a model across multiple datasets using only a few fine-tuning results, which would be significantly less expensive than performing extensive fine-tuning on numerous datasets. However, the experiments revealed that accurate predictions are achievable only for models with a substantial number of fine-tuning results, typically in the range of 10-20. Further optimization is necessary to enhance precision with fewer data points. A potential avenue to address this issue is to deepen the quality analysis of the datasets. This would help identify which datasets are most valuable for fine-tuning, thereby accelerating the optimal positioning of a new model within the embedding space.

Our method predicts the $\Delta$ value, which represents a relative position in the literature for an understudied dataset. This metric does not provide insights into the task's difficulty and is constrained by existing performance bounds (worst and best performances). Ideally, we should aim to predict the model performance directly or, at least, the absolute distance to the best performance, denoted as $\delta$. Although we have not yet achieved such a goal, future work could explore this direction.

One of the main challenges we faced while building our dataset of fine-tuning results was the discrepancies between the results of different studies, that should in theory give the same values. Many works, for example, re-implement existing methods to compare against their own approach, and in the process improve previous baselines. In some other cases, results announced in an article were not reproduced by subsequent works, despite many attempts. A more thoughtful approach to aggregating literature results than the max aggregation we used could help improve the quality of the studied data.

\section{Conclusion}

In this work, we propose a quick overview of the literature in Remote Sensing Foundation Models and remote sensing downstream datasets, and design a novel approach to efficiently benchmarking RSFMs on many datasets across various tasks. By embedding both the models and datasets in the same latent space, the proposed method allows cheap prediction of a model's fine-tuning accuracy on a given dataset. This article paves the way towards more efficient benchmarking of RSFMs without the need for extensive fine-tuning on a wide variety of downstream datasets, which is computationally expensive. Future works building on our method could enable more precise predictions based on less initial data and greatly ease the benchmarking of new RSFMs and Earth observation datasets.

{
    \small
    \bibliographystyle{ieeetr}
    \bibliography{phd}
}


\end{document}